%% file: iclr2027_conference.tex
\documentclass{article} 
\usepackage{iclr2027_conference,times}

\input{math_commands.tex}

\usepackage{hyperref}
\hypersetup{
  hidelinks,
  pdftitle={PQR3D: Progressive Query Refinement over Reference-Conditioned Temporal Windows for Multi-View 3D Object Detection},
  pdfauthor={Hui Ye, Yudong Liu, Yiran Chen, Rajshekhar Sunderraman, Shihao Ji}
}
\usepackage{url}

\usepackage{graphicx}
\usepackage{subcaption}
\usepackage{amsmath}
\usepackage[table]{xcolor}
\usepackage{pifont}
\newcommand{\cmark}{\ding{51}}

\usepackage{caption}
\usepackage{booktabs}
\usepackage{adjustbox}
\newcommand{\xmark}{\ding{55}}

\usepackage{float}
\usepackage{placeins}
\usepackage{booktabs}
\usepackage{subcaption}
\usepackage[table]{xcolor}

\title{PQR3D: Progressive Query Refinement
over \\ Reference-Conditioned Temporal Windows\\
for Multi-View 3D Object Detection}

\author{%
  Hui Ye\textsuperscript{1} \quad
  Yudong Liu\textsuperscript{2} \quad
  Yiran Chen\textsuperscript{2} \quad
  Rajshekhar Sunderraman\textsuperscript{1} \quad
  Shihao Ji\textsuperscript{3} \\
  \textsuperscript{1}Department of Computer Science, Georgia State University \\
  \textsuperscript{2}Department of Electrical and Computer Engineering, Duke University \\
  \textsuperscript{3}School of Computing, University of Connecticut \\
    \texttt{\{hye2,rsunderraman\}@gsu.edu \quad shihao.ji@uconn.edu} \\
  \texttt{\{yudong.liu,yiran.chen\}@duke.edu}
}

\iclrfinalcopy 
\begin{document}

\maketitle
\lhead{Preprint}

\begin{abstract}
Temporal context is essential for camera-only multi-view 3D object detection. Existing streaming detectors maintain and propagate query states from one frame to the next, requiring sequence-aware training and chronological inference. We propose PQR3D, which performs progressive query refinement within reference-conditioned temporal windows. This design enables random frame sampling and independent inference without persistent query memory. Within each window, PQR3D progressively transfers motion-aligned high-confidence queries from earlier timestamps toward the target frame. We further introduce masked self-attention to regulate interactions among regular, propagated, and denoising queries while keeping denoising supervision isolated from detection queries. In addition, a stage-decoupled anchor embedding injects position before self-attention and size, orientation, and velocity afterward, reducing interference from temporally inconsistent attributes.
 With a ViT-L backbone, PQR3D sets a new state of the art on the nuScenes test set, achieving 71.6 NDS and 64.9 mAP. Source code is available at \url{https://github.com/huiyegit/PQR3D}

\end{abstract}

\section{Introduction}

3D perception is a fundamental task in autonomous vehicles, robotics, virtual reality, and augmented reality. Multiple sensors, including cameras, LiDAR, and radar, can be leveraged to improve perception performance. Among them, cameras capture rich semantic cues, such as object boundaries, textures, and background context, which closely align with the human visual system.

Recent query-based camera-only methods use 3D reference points to aggregate image features for 3D object detection. With temporal modeling, they achieve strong accuracy and efficiency. Existing temporal query-based methods mainly follow two strategies. The first directly samples and aggregates image features from multiple timestamps for each current-frame query~\cite{lin2022sparse4d,liu2023sparsebev}. However, it fuses past features directly rather than progressively refining queries across frames. The second sequentially propagates cached query states between video frames. StreamPETR~\cite{wang2023exploring} stores high-confidence object queries in a memory queue and propagates them frame by frame, while Sparse4Dv2 and Sparse4Dv3~\cite{lin2023sparse4dv2,lin2023sparse4dv3} transfer instance features and structured anchors from the preceding frame to the current frame. These methods effectively model long-term temporal information but require chronological processing because each prediction depends on states from earlier frames.

\begin{figure*}[t!]
\centering
\includegraphics[width=1.0\textwidth]{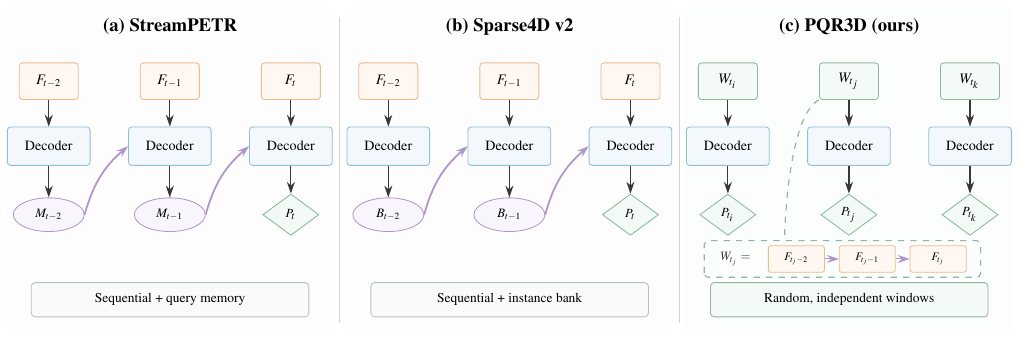}
\vspace{-20pt}
\caption{
\textbf{Comparison of temporal propagation strategies.}
(a) StreamPETR processes frames sequentially and passes query memory $M_t$ from one frame to the next.
(b) Sparse4D v2 performs sequential propagation by passing the instance bank $B_t$ from one frame to the next.
(c) PQR3D processes each reference-conditioned temporal window independently, without carrying memory across target samples. The dashed box shows one window, $W_{t_j}=\left\{F_{t_j-2},F_{t_j-1},F_{t_j}\right\}$, whose features are processed in temporal order to produce the 3D detection result $P_{t_j}$ for the target frame. Here, $F_t$ and $P_t$ denote the multi-view image features and 3D detection result at time $t$, respectively.
}
\label{fig:propagation}
\vspace{-15pt}
\end{figure*}

However, sequential query propagation with persistent memory introduces several limitations.
 First, the model must preserve scene boundaries and chronological order during training and inference, preventing target frames from being processed as fully independent samples. This requirement limits random target-frame sampling during training and prevents frames from being processed independently during inference. Second, inaccurate queries from earlier frames can be propagated forward and affect predictions in later frames. Finally, these methods must maintain a temporal memory and reset it whenever the input video sequence changes.
 As illustrated in Figure~\ref{fig:propagation}, both StreamPETR and Sparse4D v2 maintain persistent states across successive frames.

To address these limitations, we propose PQR3D, a framework for progressive query refinement over reference-conditioned temporal windows. As shown in Figure~\ref{fig:propagation}, PQR3D constructs a self-contained window, such as $W_{t_j}=\{F_{t_j-2}, F_{t_j-1}, F_{t_j}\}$, for each target frame. Within each window, object queries are progressively refined from earlier timestamps toward the target timestamp. At each step, temporal features are aligned to the current frame, and high-confidence queries are motion-compensated before being passed to the next step.
 After predicting the target frame, PQR3D discards the local state, allowing each temporal window to be processed independently.

In addition, to combine query denoising with propagated queries within each temporal window, we introduce masked self-attention that isolates denoising queries from regular and propagated queries while allowing full interaction between the latter two groups.
Furthermore, Sparse4D v3 separately encodes multiple anchor attributes, including position, size, orientation, and velocity, and incorporates the resulting representation into attention~\cite{lin2023sparse4dv3}. However, directly injecting the complete anchor embedding before self-attention degrades detection performance in PQR3D, where regular and propagated queries interact within the same temporal window. We therefore propose a stage-decoupled anchor embedding (SDAE), which injects different anchor attributes at different stages of the decoder. The position embedding is added before self-attention to guide spatial interactions, whereas the size, orientation, and velocity embeddings are injected afterward to refine object geometry and motion.

Our main contributions are summarized as follows:
1) We present PQR3D, a progressive query refinement framework that processes each reference-conditioned temporal window independently and refines object queries in temporal order toward the target frame, without carrying memory across target samples.
2) We introduce a masked self-attention mechanism tailored to progressive query refinement, which isolates denoising queries from regular and propagated queries while preserving temporal interactions between the latter two.
3) We propose a stage-decoupled anchor embedding that injects position before self-attention and size, orientation, and velocity afterward, separating spatial query interaction from geometric and motion refinement.
4) PQR3D achieves state-of-the-art performance on the nuScenes benchmark. It consistently outperforms existing methods across different backbones, while requiring fewer training epochs.

\section{Related work}
\label{gen_inst}


\subsection{BEV-based  methods}

BEV-based methods explicitly predict dense depth to lift 2D image features into 3D space, which are then projected onto the BEV plane to create BEV grid features.  
BEVDet~\cite{huang2021bevdet} builds upon the LSS~\cite{philion2020lift} paradigm and introduces a BEV-based 3D detection framework with specifically designed data augmentation and NMS strategies. 
BEVDepth~\cite{li2022bevdepth} enhances the LSS paradigm with explicit depth supervision from LiDAR point clouds and camera-aware depth prediction, improving depth estimation quality.
AeDet~\cite{feng2022aedet} proposes an azimuth-equivariant convolution (AeConv) and an azimuth-equivariant anchor to preserve the inherent radial symmetry of BEV features and facilitate detector optimization.

With advanced temporal modeling that leverages multi-frame BEV features, recent BEV-based methods have achieved superior performance over single-frame methods.
BEVDet4D~\cite{huang2022bevdet4d} extends BEVDet with temporal fusion by aligning the previous frame's BEV feature to the current frame via ego-motion compensation and concatenating the two. 
BEVStereo~\cite{li2023bevstereo} improves depth estimation by introducing temporal multi-view stereo that utilizes adjacent frames for stereo matching, with dynamic candidate selection to reduce computational cost.
BEVFormer series~\cite{li2022bevformer,yang2023bevformer} propose unified BEV representations learned via spatiotemporal transformers, using spatial cross-attention to aggregate multi-view features into predefined BEV queries and temporal self-attention to recurrently fuse historical BEV features.
SOLOFusion~\cite{park2023time} combines long-term, coarse stereo matching with short-term, fine-grained matching, showing that the two are complementary for camera-based 3D detection.
HoP~\cite{zong2023temporal} introduces an auxiliary task that predicts objects at a historical timestamp $t\!-\!k$ from a pseudo BEV feature, improving spatial and temporal representations without inference overhead.

\subsection{Sparse query-based methods}

Query-based methods initialize 3D reference points within the detection range and use them to aggregate image features. DETR3D~\cite{wang2022detr3d} uses sparse 3D object queries to sample multi-view 2D features through camera transformations. The PETR series~\cite{liu2022petr,liu2023petrv2} encodes 3D positional information into image features, allowing object queries to interact directly with position-aware representations for end-to-end detection.

Recent query-based methods have substantially improved detection performance by incorporating temporal modeling over multi-frame image features.
The Sparse4D series~\cite{lin2022sparse4d, lin2023sparse4dv2, lin2023sparse4dv3} progressively improves sparse query-based detection. Sparse4D samples multi-view, multi-scale, multi-frame features through projected 3D keypoints. Sparse4Dv2 propagates sparse features recurrently for long-term modeling at reduced cost. Sparse4Dv3 adds temporal instance denoising, quality estimation, and decoupled attention.
StreamPETR~\cite{wang2023exploring} proposes an object-centric temporal mechanism that propagates historical information through object queries frame by frame, along with a motion-aware layer normalization to capture object motion.
Building upon StreamPETR, RayDN~\cite{liu2024ray} introduces ray denoising, which samples hard negatives along camera rays based on depth distributions, encouraging the model to learn depth-aware features for more accurate detection.
SparseBEV~\cite{liu2023sparsebev} introduces query-guided adaptive spatio-temporal sampling and adaptive mixing, which decode sampled features through dynamic weights derived from the queries.
Far3D~\cite{jiang2024far3d} leverages high-quality 2D object priors to generate 3D adaptive queries that complement the 3D global queries, and introduces a perspective-aware aggregation module to effectively aggregate multi-view, multi-scale features for long-range detection.
The HENet series~\cite{xia2024henet, xia2025henet} introduces a hybrid image encoding strategy, applying a lightweight encoder to long-term frames and a large encoder to short-term frames, with an attention-based module to integrate temporal features from both encoders. In contrast to direct multi-frame feature aggregation and persistent memory sequential propagation, PQR3D performs progressive query refinement within reference-conditioned temporal windows, each constructed and processed independently. Within each window, PQR3D aligns temporal features and progressively refines queries toward the target frame. The local state is discarded after prediction, enabling random target-frame sampling during training and independent inference without persistent memory.

\begin{figure*}[t!]
\begin{center}
\includegraphics[width=1.0\textwidth]{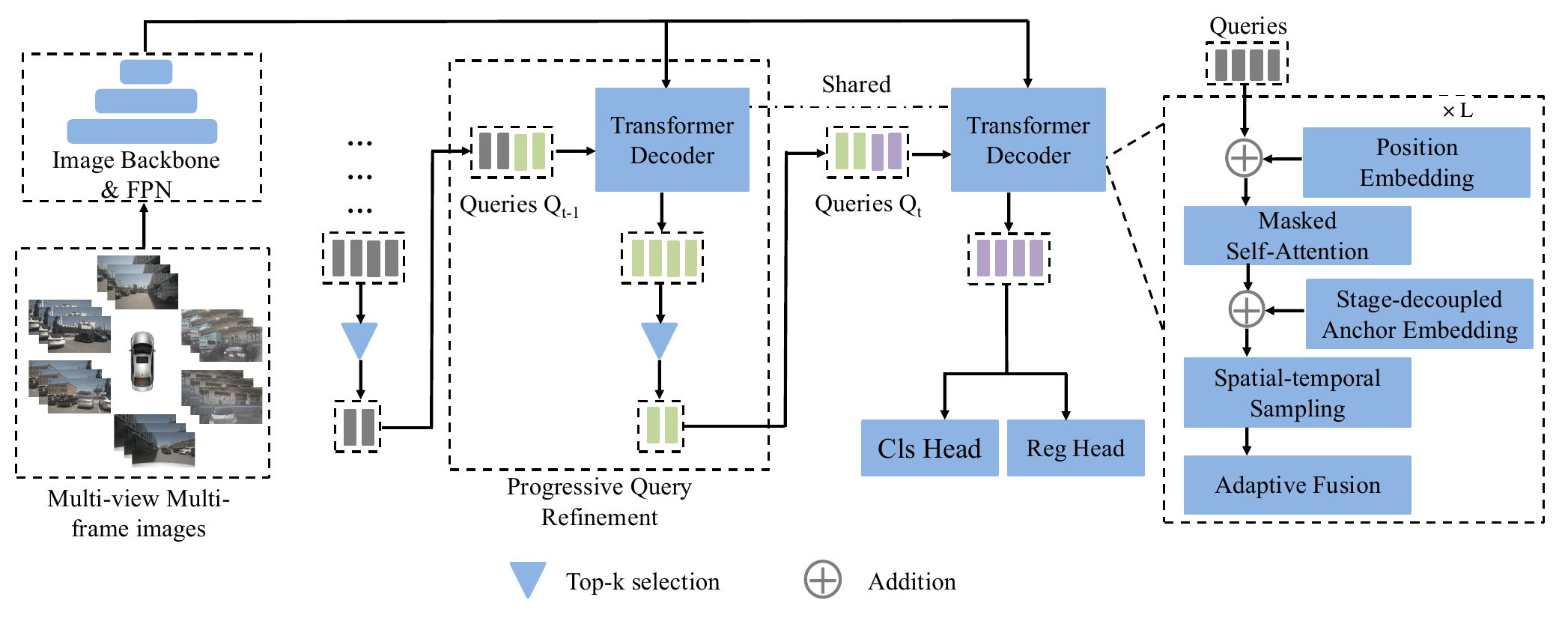}
\end{center}

\vspace{-10pt}
\caption{\textbf{Overall architecture of PQR3D.}
Within each temporal window, high-confidence queries are propagated from earlier frames toward the current frame, concatenated with the current queries $Q_t$, and progressively refined by Transformer decoders. Each decoder contains $L$ layers (right), consisting of position embedding, masked self-attention, stage-decoupled anchor embedding, spatio-temporal sampling, and adaptive fusion.}

\label{fig:architecture}
 \vspace{-10pt}
\end{figure*}

\section{Method}
\label{method}


As shown in Fig.~\ref{fig:architecture}, PQR3D follows an encoder-decoder 
architecture with progressive query refinement.
Given multi-view multi-frame images at timestamp $t$, an image encoder 
(e.g., ResNet-50) with FPN extracts the feature maps to form the 
multi-frame feature sequence $\mathcal{F}_t = \{\mathbf{F}_t, 
\mathbf{F}_{t-1}, \cdots, \mathbf{F}_{t-N}\}$, where $\mathbf{F}_t$ 
denotes the multi-view multi-scale feature at timestamp $t$.
A set of learnable object queries  $\mathbf{Q}_{t}$ are initialized in  the detection range of BEV space, where each query corresponds to a grid cell region on the BEV plane.
The object queries $\mathbf{Q}_t$ and image features $\mathcal{F}_t$ are fed into a Transformer decoder $\mathcal{D}$ containing $L$ layers. Each layer comprises position embedding, masked self-attention, stage-decoupled anchor embedding, spatial-temporal sampling, and adaptive fusion.

\subsection{Progressive Query Refinement}
Given the multi-frame image feature sequence 
$\mathcal{F}_t = \{\mathbf{F}_{t}, \mathbf{F}_{t-1}, \cdots, \mathbf{F}_{t-N}\}$ 
consisting of the current and $N$ historical frame image features, 
the projection matrix sequence 
$\{\mathbf{P}_{t}, \mathbf{P}_{t-1}, \cdots, \mathbf{P}_{t-N}\}$ 
which projects coordinates from 3D space to the image plane, 
and the object queries $\{\mathbf{Q}_{t}, \mathbf{Q}_{t-1}, \cdots, \mathbf{Q}_{t-N}\}$, 
where $t-i$ denotes the timestamp and $i \in \{0, 1, \cdots, N\}$,
the progressive query refinement is formulated as follows. 

At keyframe $t-i$ ($0 < i \leq N$), the image feature sequence 
$\mathcal{F}_t$ is reordered by circularly shifting so that 
$\mathbf{F}_{t-i}$ is placed at the front, producing the reordered 
image feature sequence 
$\hat{\mathcal{F}}_{t-i} = \{\mathbf{F}_{t-i}, \mathbf{F}_{t-i-1}, \cdots, 
\mathbf{F}_{t-N}, \mathbf{F}_{t}, \cdots, \mathbf{F}_{t-i+1}\}$. 
This reordering ensures that the target keyframe $t-i$ serves as the reference frame, while treating all other keyframes as temporal context.
The transformer decoder $\mathcal{D}$ 
takes the reordered image feature sequence $\hat{\mathcal{F}}_{t-i}$, 
projection matrix $\mathbf{P}_{t-i}$, and object queries 
$\mathbf{Q}_{t-i}$ as input to produce the refined queries:
\begin{equation}
\mathbf{Q}'_{t-i} = \mathcal{D}(\mathbf{Q}_{t-i}, \hat{\mathcal{F}}_{t-i}, \mathbf{P}_{t-i}).
\end{equation}

The refined queries $\mathbf{Q}'_{t-i}$, which encode the predicted 
bounding box attributes (position, size, orientation, and velocity), 
are then transformed from the coordinate system at timestamp $t-i$ 
to the next timestamp $t-i+1$. Specifically, given the ego-motion 
matrix $\mathbf{E}_{t-i \rightarrow t-i+1}$ between the two timestamps 
and the time interval $\Delta t$, we first rotate the velocity by 
the 2D rotation component $\mathbf{E}^{r}_{t-i \rightarrow t-i+1}$ 
of $\mathbf{E}_{t-i \rightarrow t-i+1}$:
\begin{equation}
\tilde{\mathbf{v}}_{t-i} = {\mathbf{E}^{r}_{t-i \rightarrow t-i+1}}^{\top} \, \mathbf{v}_{t-i}.
\end{equation}
Then the 3D position $\mathbf{p}_{t-i}$ is transformed by 
$\mathbf{E}_{t-i \rightarrow t-i+1}$ with velocity compensation:
\begin{equation}
\tilde{\mathbf{p}}_{t-i} = \mathbf{E}_{t-i \rightarrow t-i+1} \, 
\mathbf{p}_{t-i} + \tilde{\mathbf{v}}_{t-i} \cdot \Delta t,
\end{equation}

where $\tilde{\mathbf{p}}_{t-i}$ is the compensated position in 
the coordinate system of timestamp $t-i+1$. Together with the 
rotated velocity and other bounding box attributes, they form 
the transformed queries $\tilde{\mathbf{Q}}_{t-i}$ in the next 
keyframe's coordinate system.


After the coordinate transformation, we select the top-$k$ queries 
with the highest classification confidence scores from the 
transformed queries. This filtering step retains only the most 
reliable predictions for propagation, reducing noise from 
low-confidence detections:
\begin{equation}
\hat{\mathbf{Q}}_{t-i} = \mathrm{TopK}(\tilde{\mathbf{Q}}_{t-i}).
\end{equation}

The selected queries $\hat{\mathbf{Q}}_{t-i}$ are concatenated with 
the object queries $\mathbf{Q}_{t-i+1}$ at the next keyframe to 
form the input queries for the next propagation step:
\begin{equation}
\bar{\mathbf{Q}}_{t-i+1} = \hat{\mathbf{Q}}_{t-i} \| \mathbf{Q}_{t-i+1},
\end{equation}
where $\|$ denotes concatenation.
The concatenated queries 
are then fed into the transformer decoder:
\begin{equation}
\mathbf{Q}'_{t-i+1} = \mathcal{D}(\bar{\mathbf{Q}}_{t-i+1}, 
\hat{\mathcal{F}}_{t-i+1}, \mathbf{P}_{t-i+1}).
\end{equation}
This process repeats recurrently until reaching the current 
timestamp $t$, where the final refined queries $\mathbf{Q}'_{t}$ 
are used for object classification and bounding box regression. 


\subsection{Masked Self-Attention}


\textbf{Denoising queries.}
Following DN-DETR~\cite{li2022dn}, we construct $G$ groups of denoising queries by perturbing ground-truth boxes and labels. For each box, the center offsets satisfy $|\Delta x|<\lambda w/2$ and $|\Delta y|<\lambda l/2$, where $\lambda$ controls the noise scale; we also randomly replace a fraction $\gamma$ of the class labels.




\textbf{Attention mask.} To prevent information leakage while allowing regular and propagated queries to attend to each other, we design a masked attention mechanism that controls interactions among the three query types. 
Given denoising queries $\mathbf{Q}_{dn}$, regular queries $\mathbf{Q}_t$, and propagated queries $\hat{\mathbf{Q}}_{t-1}$, the decoder input is
$\mathbf{Q}^{*}=\mathbf{Q}_{dn}\|\mathbf{Q}_{t}\|\hat{\mathbf{Q}}_{t-1}$,
where $\|$ denotes concatenation.
The attention mask  $\mathbf{M}\in\{0,-\infty\}^{W\times W}$ is defined as:
\begin{equation}
\mathbf{M}_{ij} =
\begin{cases}
0, & i,j\in\mathbf{Q}_{dn}^{g}
     \text{ for the same group }g,\\
0, & i,j\in\mathbf{Q}_{t}\cup\hat{\mathbf{Q}}_{t-1},\\
-\infty, & \text{otherwise},
\end{cases}
\end{equation}
where $W$ is the total number of queries and $\mathbf{M}_{ij} = -\infty$ indicates that query $i$ is blocked from attending to query $j$. This design ensures that: (1) denoising queries attend only within their own group, preventing cross-group label leakage; (2) regular and propagated queries attend freely to each other, enabling temporal information exchange; and (3) denoising queries are isolated from regular and propagated queries, separating the auxiliary denoising task from the main detection task.

Before the masked self-attention, the masked position embedding $\bar{\mathbf{E}}_{\mathrm{pos}}$ of our stage-decoupled anchor embedding (described in Sec.~\ref{sec:dae}) is added to the queries to encode spatial information:
\begin{equation}
\tilde{\mathbf{Q}}^{*} = \mathbf{Q}^{*} + \bar{\mathbf{E}}_{\mathrm{pos}}.
\label{eq:pos_embed}
\end{equation}

Given the pairwise BEV distance matrix $\mathbf{D} \in \mathbb{R}^{W \times W}$, where $\mathbf{D}_{ij} = \sqrt{(x_i - x_j)^2 + (y_i - y_j)^2}$ is the BEV distance between the centers of queries $i$ and $j$, the masked self-attention is computed as:
\begin{equation}
\mathrm{Attn}(\tilde{\mathbf Q}^{*})
=
\mathrm{softmax}\!\left(
\frac{\tilde{\mathbf Q}^{*}(\tilde{\mathbf Q}^{*})^\top}{\sqrt d}
-\tau\mathbf D+\mathbf M
\right)\tilde{\mathbf Q}^{*},
\end{equation}
where $d$ is the feature dimension and $\tau$ is a learnable coefficient that controls the receptive field. 

\begin{figure*}[t!]
\begin{center}
\includegraphics[width=1.0\textwidth]{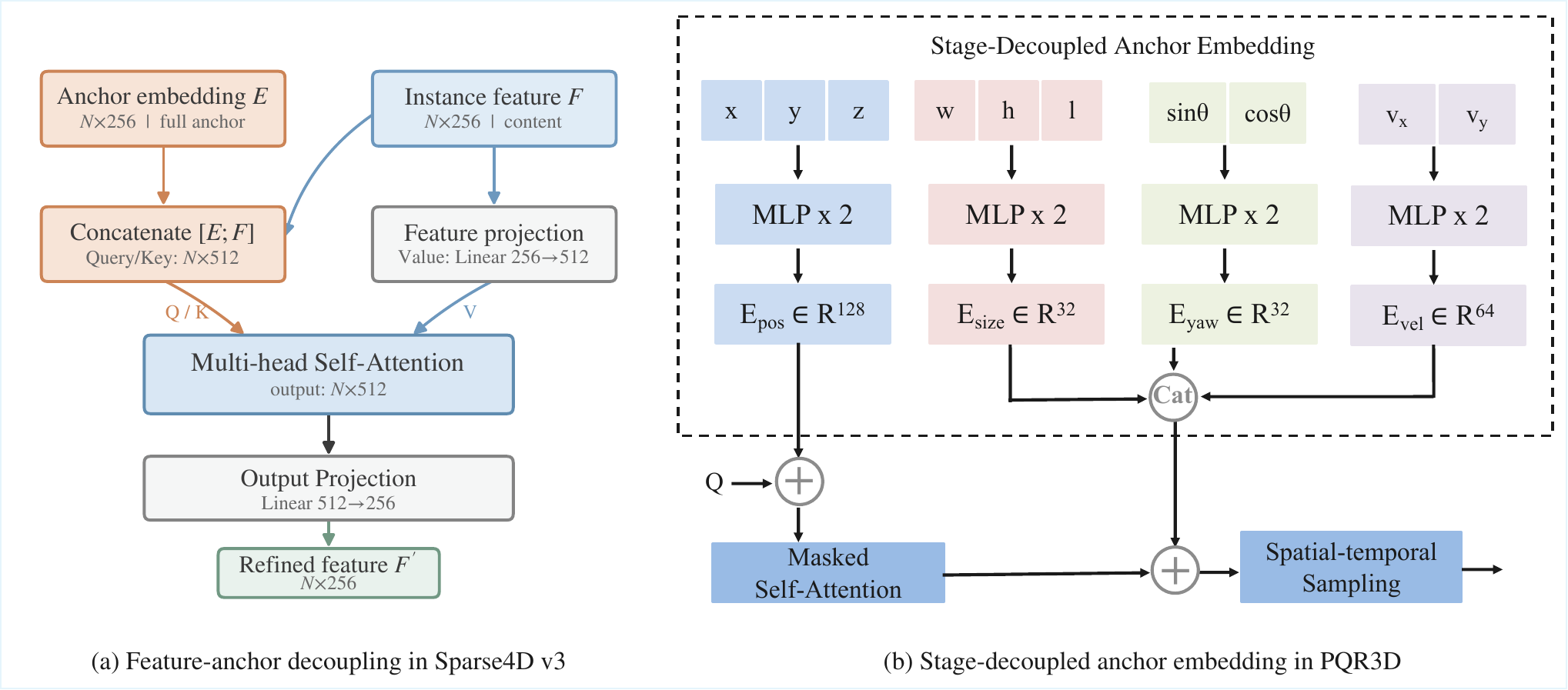}
\end{center}
\vspace{-5pt}
\caption{
\textbf{Comparison of anchor embedding strategies.}
(a) Sparse4D v3 constructs the query and key by concatenating the anchor embedding $E$ with the instance feature $F$, while deriving the value from $F$ alone.
(b) Our stage-decoupled anchor embedding injects the position embedding $E_{\mathrm{pos}}$ before masked self-attention, while the size, yaw, and velocity embeddings are concatenated and injected afterward, before spatio-temporal sampling.
$\oplus$ denotes element-wise addition, and \textit{Cat} denotes concatenation.
}
\label{fig:sdae}
 \vspace{-10pt}
\end{figure*}

\subsection{Stage-Decoupled Anchor Embedding}
\label{sec:dae}

Several query-based detectors~\cite{liu2023sparsebev,wang2023exploring,liu2024ray}
primarily condition object queries on position embeddings derived from the
anchor center $(x,y,z)$. Sparse4Dv3~\cite{lin2023sparse4dv3} further encodes
the full anchor and introduces feature--anchor decoupling. As shown in
Fig.~\ref{fig:sdae}(a), it concatenates the anchor embedding $\mathbf{E}$
with the instance feature $\mathbf{F}$ to construct the query and key, while
using the projected instance feature as the value. Thus, all anchor attributes
are jointly introduced before self-attention.

In contrast, our stage-decoupled anchor embedding injects different anchor
attributes at different decoder stages, as illustrated in
Fig.~\ref{fig:sdae}(b). Position is introduced before masked self-attention,
whereas size, orientation, and velocity are introduced afterward. Specifically, given an anchor
\begin{equation}
\mathbf{a}
=
(x,y,z,w,h,l,\sin\theta,\cos\theta,v_x,v_y),
\end{equation}
we encode its attribute groups using separate MLPs:
\begin{align}
\mathbf{E}_{\mathrm{pos}}
&=
\mathrm{MLP}_{\mathrm{pos}}(x,y,z)
\in\mathbb{R}^{128},
&
\mathbf{E}_{\mathrm{size}}
&=
\mathrm{MLP}_{\mathrm{size}}(w,h,l)
\in\mathbb{R}^{32},
\nonumber\\
\mathbf{E}_{\mathrm{yaw}}
&=
\mathrm{MLP}_{\mathrm{yaw}}(\sin\theta,\cos\theta)
\in\mathbb{R}^{32},
&
\mathbf{E}_{\mathrm{vel}}
&=
\mathrm{MLP}_{\mathrm{vel}}(v_x,v_y)
\in\mathbb{R}^{64}.
\end{align}
Each MLP consists of two linear layers with ReLU activations. We concatenate
the size, orientation, and velocity embeddings to form
\begin{equation}
\mathbf{E}_{\mathrm{attr}}
=
\left[
\mathbf{E}_{\mathrm{size}}
\;\|\;
\mathbf{E}_{\mathrm{yaw}}
\;\|\;
\mathbf{E}_{\mathrm{vel}}
\right]
\in\mathbb{R}^{128}.
\label{eq:attr_concat}
\end{equation}
The complete anchor embedding is
\begin{equation}
\mathbf{E}_{\mathrm{anchor}}
=
\left[
\mathbf{E}_{\mathrm{pos}}
\;\|\;
\mathbf{E}_{\mathrm{attr}}
\right]
\in\mathbb{R}^{256}.
\end{equation}

We then apply channel masking to obtain two 256-dimensional embeddings:
\begin{equation}
\bar{\mathbf{E}}_{\mathrm{pos}}
=
\left[
\mathbf{E}_{\mathrm{pos}}
\;\|\;
\mathbf{0}_{128}
\right],
\qquad
\bar{\mathbf{E}}_{\mathrm{attr}}
=
\left[
\mathbf{0}_{128}
\;\|\;
\mathbf{E}_{\mathrm{attr}}
\right].
\label{eq:masked_anchor_embeddings}
\end{equation}
For $\bar{\mathbf{E}}_{\mathrm{pos}}$, the attribute channels are masked out;
for $\bar{\mathbf{E}}_{\mathrm{attr}}$, the position channels are masked out.
Both embeddings remain 256-dimensional and are therefore compatible with
the query features.
The masked position embedding $\bar{\mathbf{E}}_{\mathrm{pos}}$ is added to the queries before self-attention, as in Eq.~(\ref{eq:pos_embed}).
After masked self-attention, the masked attribute embedding is added as a
residual:
\begin{equation}
\mathbf{Q}'
=
\mathrm{Attn}(\tilde{\mathbf{Q}}^{*})
+
\bar{\mathbf{E}}_{\mathrm{attr}}.
\label{eq:attr_residual}
\end{equation}

This stage-decoupled design uses position to guide query interactions while
introducing geometric and motion attributes afterward, reducing interference
from inconsistent attributes between regular and propagated queries.

\noindent\textbf{Spatial-temporal Sampling and Adaptive Fusion.}
After masked self-attention and stage-decoupled anchor embedding, we follow SparseBEV~\cite{liu2023sparsebev} to aggregate multi-view multi-frame image features. The spatial-temporal sampling module generates adaptive 3D sampling offsets from query features, warps them across timestamps using ego-motion and object velocity, and projects them onto multi-view feature maps for bilinear sampling. The adaptive fusion module then fuses the sampled features via dynamic channel and point mixing. Finally, the updated queries are passed through a feed-forward network and decoded by two separate MLP heads for classification and box regression.

\section{Experiments}
\label{experiment}

 \subsection{Implementation Details}
 \label{subsec:implementation_details}

\noindent\textbf{BEV Flip Augmentation.}
We introduce a BEV flip augmentation designed for progressive query refinement. 
During training, we apply a mutually exclusive random flip along either the $x$-axis or the $y$-axis to the BEV coordinate system.
We apply the flip only to the current frame's projection and ego-pose matrices, while keeping historical frames in their original coordinate systems. Through the relative ego-pose transformation, the inverse of the flipped ego-pose at $t$ naturally maps past queries into the flipped coordinate system.
This keeps current-frame and propagated historical queries in a consistent coordinate system, enabling stable training.

We conduct experiments with ResNet-50, ResNet-101~\cite{he2016deep}, 
VoV99~\cite{lee2019energy}, and ViT-L~\cite{dosovitskiy2020image} backbones. 
ResNet-50 and ResNet-101 are initialized using ImageNet~\cite{deng2009imagenet} 
and nuImages~\cite{caesar2020nuscenes} pre-training and evaluated on the 
nuScenes validation set. VoV99 is initialized from DD3D~\cite{park2021pseudo}, 
while VoV99 and ViT-L results are reported on the test set.


\subsection{Dataset and Metrics}
\label{subsec:dataset_metrics}

We evaluate PQR3D on nuScenes~\cite{caesar2020nuscenes}, a large-scale autonomous driving benchmark with $1000$ driving scenes captured by $6$ surround-view cameras, $1$ LiDAR, and $5$ radars.
The dataset is split into $700$/$150$/$150$ scenes for training, validation, and testing.  Each scene spans $20$ seconds with key samples annotated at $2$ Hz, providing $1.4$M annotated 3D bounding boxes across $10$ object classes.
We report the nuScenes Detection Score (NDS) as the primary metric, which is a weighted combination of mean Average Precision (mAP) and five True Positive (TP) metrics: average translation error (ATE), average scale error (ASE), average orientation error (AOE), average velocity error (AVE), and average attribute error (AAE). NDS is a weighted combination of mAP and the five TP metrics (see~\cite{caesar2020nuscenes} for the precise formulation), providing a holistic evaluation. 

\begin{table}[t]
\centering
\caption{Performance comparison on the nuScenes val split. $\ddagger$ indicates methods with CBGS which elongates 1 epoch into 4.5 epochs. 
$\ast$ denotes our reproduction at 24 epochs using ResNet-101 
configurations adapted from the official ResNet-50 release, since neither RayDN nor 
Sparse4Dv3 provides ResNet-101 configurations.
}
\label{tab:val}
\scriptsize
\begin{adjustbox}{width=\textwidth}
\begin{tabular}{l|ccc|cc|ccccc}
\toprule
Method & Backbone & Input Size & Epochs & NDS$\uparrow$ & mAP$\uparrow$ & mATE$\downarrow$ & mASE$\downarrow$ & mAOE$\downarrow$ & mAVE$\downarrow$ & mAAE$\downarrow$ \\
\midrule
PETRv2~\cite{liu2023petrv2} & ResNet50 & $800\times320$ & 60 & 45.6 & 35.0 & 0.726 & 0.277 & 0.505 & 0.503 & 0.181 \\
BEVStereo~\cite{li2023bevstereo} & ResNet50 & $704\times256$ & 90$\ddagger$ & 50.0 & 37.2 & 0.598 & 0.270 & 0.438 & 0.367 & 0.190 \\
BEVPoolv2~\cite{huang2022bevpoolv2} & ResNet50 & $704\times256$ & 90$\ddagger$ & 52.6 & 40.6 & 0.572 & 0.275 & 0.463 & 0.275 & 0.188 \\
SOLOFusion~\cite{park2023time} & ResNet50 & $704\times256$ & 90$\ddagger$ & 53.4 & 42.7 & 0.567 & 0.274 & 0.511 & 0.252 & 0.181 \\
Sparse4Dv2~\cite{lin2023sparse4dv2} & ResNet50 & $704\times256$ & 100 & 53.9 & 43.9 & 0.598 & 0.270 & 0.475 & 0.282 & 0.179 \\
StreamPETR~\cite{wang2023exploring} & ResNet50 & $704\times256$ & 60 & 55.0 & 45.0 & 0.613 & 0.267 & 0.413 & 0.265 & 0.196 \\
SparseBEV~\cite{liu2023sparsebev} & ResNet50 & $704\times256$ & 36 & 55.8 & 44.8 & 0.581 & 0.271 & 0.373 & 0.247 & 0.190 \\
Sparse4Dv3~\cite{lin2023sparse4dv3} & ResNet50 & $704\times256$ & 100 & 56.1 &  \underline{46.9} & 0.553 & 0.274 & 0.476 & 0.227 & 0.200 \\
RayDN~\cite{liu2024ray} & ResNet50 & $704\times256$ & 60 & \underline{56.3} & \underline{46.9} & 0.579 & 0.264 & 0.433 & 0.256 & 0.187 \\
\rowcolor{gray!15}
\textbf{PQR3D}
& ResNet50
& $704\times256$
& 24
& \textbf{58.2}
& \textbf{48.6}
& 0.576
& 0.263
& 0.353
& 0.224
& 0.191 \\
\midrule
DETR3D~\cite{wang2022detr3d} & ResNet101-DCN & $1600\times900$ & 24 & 43.4 & 34.9 & 0.716 & 0.268 & 0.379 & 0.842 & 0.200 \\
BEVFormer~\cite{li2022bevformer} & ResNet101-DCN & $1600\times900$ & 24 & 51.7 & 41.6 & 0.673 & 0.274 & 0.372 & 0.394 & 0.198 \\
BEVDepth~\cite{li2022bevdepth} & ResNet101 & $1408\times512$ & 90$\ddagger$ & 53.5 & 41.2 & 0.565 & 0.266 & 0.358 & 0.331 & 0.190 \\

HoP-BEVFormer~\cite{zong2023temporal} & ResNet101-DCN & $1600\times900$ & 24 & 55.8 & 45.4 & 0.565 & 0.265 & 0.327 & 0.337 & 0.194 \\
SOLOFusion~\cite{park2023time} & ResNet101 & $1408\times512$ & 90$\ddagger$ & 58.2 & 48.3 & 0.503 & 0.264 & 0.381 & 0.246 & 0.207 \\

Sparse4Dv2~\cite{lin2023sparse4dv2} & ResNet101 & $1408\times512$ & 100  & \underline{59.4} & \underline{50.5} & 0.548 & 0.268 & 0.348 & 0.239 & 0.184 \\

SparseBEV~\cite{liu2023sparsebev} & ResNet101 & $1408\times512$ & 24 & 59.2 & 50.1 & 0.562 & 0.265 & 0.321 & 0.243 & 0.195 \\

Sparse4Dv3$^\ast$~\cite{lin2023sparse4dv3} & ResNet101 & $1408\times512$ & 24 & 56.1 & 48.0 & 0.556 & 0.261 & 0.535 & 0.237 & 0.200 \\

RayDN$^\ast$~\cite{liu2024ray} & ResNet101 & $1408\times512$ & 24 & 55.1 & 46.0 & 0.610 & 0.270 & 0.431 & 0.279 & 0.204 \\

\rowcolor{gray!15}
\textbf{PQR3D}
& ResNet101
& $1408\times512$
& 24
& \textbf{61.2}
& \textbf{52.0}
& 0.540
& 0.261
& 0.276
& 0.214
& 0.196 \\

\bottomrule
\end{tabular}
\end{adjustbox}
\vspace{-5pt}
\end{table}

\subsection{Main Results}

\textbf{Validation set.} As shown in Table~\ref{tab:val}, we compare PQR3D with existing methods on the nuScenes val split. With ResNet-50 as the backbone and an input resolution of $704 \times 256$, PQR3D achieves $58.2$ NDS and $48.6$ mAP, surpassing the previous best method RayDN by $+1.9$ NDS and $+1.7$ mAP. Notably, PQR3D is trained for only 24 epochs, while RayDN requires 60 epochs to reach its reported performance. Compared with SparseBEV, which uses a relatively short 36-epoch schedule, PQR3D achieves further gains of $+2.4$ NDS and $+3.8$ mAP with only 24 training epochs. These results demonstrate that PQR3D consistently outperforms prior state-of-the-art methods under the ResNet-50 setting while requiring  fewer training epochs. 

When scaling to ResNet-101, RayDN and Sparse4Dv3 report results from models trained for 60 and 100 epochs, respectively, but do not release the corresponding configuration files.
 For a fair comparison
under the 24-epoch setting, we adapt their open-source ResNet-50 configurations to ResNet-101
and report the reproduced results.
As shown in Table~\ref{tab:val}, PQR3D achieves an NDS of $61.2$ and an mAP of $52.0$,
outperforming the previous best model Sparse4Dv2, by $+1.8$ NDS and $+1.5$ mAP,
despite being trained for only 24 epochs, whereas Sparse4Dv2 is trained for 100 epochs.
Compared with SparseBEV, which uses the same 24-epoch setting, PQR3D further improves
performance by $+2.0$ NDS and $+1.9$ mAP. These results demonstrate that PQR3D scales effectively to larger backbones and higher input resolutions, while maintaining strong training efficiency.

\textbf{Test set.} As shown in Table~\ref{tab:test}, we compare PQR3D with existing camera-only methods on the nuScenes test split. With VoV99, PQR3D achieves state-of-the-art performance among methods using the same backbone, reaching 68.0 NDS and 61.0 mAP. Under the same $1600\times640$ input resolution and 24-epoch training schedule, PQR3D outperforms SparseBEV by 0.5 NDS and 0.7 mAP points. With ViT-L, PQR3D achieves 71.6 NDS and 64.9 mAP, establishing new state-of-the-art performance. Compared with SparseBEV under the same backbone, input resolution, and training schedule, PQR3D improves both NDS and mAP by 1.4 points. It also surpasses RoPETR-e by 0.7 NDS and 0.1 mAP, despite RoPETR-e using a higher $1600\times900$ resolution and test-time augmentation.





\begin{table*}[t]
\centering
\caption{
Performance comparison on the nuScenes test split.
$\dagger$ indicates configurations that use future frames.
$\ddagger$ indicates training with CBGS, for which
one nominal epoch corresponds to approximately 4.5 standard epochs. $\ast$ indicates the use of test-time augmentation.
}
\label{tab:test}
\scriptsize
\begin{adjustbox}{width=\textwidth}
\begin{tabular}{l|ccc|cc|ccccc}
\toprule
Method
& Backbone
& Input Size
& Epochs
& NDS$\uparrow$
& mAP$\uparrow$
& mATE$\downarrow$
& mASE$\downarrow$
& mAOE$\downarrow$
& mAVE$\downarrow$
& mAAE$\downarrow$ \\
\midrule

PETRv2~\cite{liu2023petrv2}
& VoV99 & $1600\times640$ & 24
& 58.2 & 49.0
& 0.561 & 0.243 & 0.361 & 0.343 & 0.120 \\

Sparse4D~\cite{lin2022sparse4d}
& VoV99 & $1600\times640$ & 48
& 59.5 & 51.1
& 0.533 & 0.263 & 0.369 & 0.317 & 0.124 \\

MV2D-T~\cite{wang2023object}
& VoV99 & $1600\times640$ & 72
& 59.6 & 51.1
& 0.525 & 0.243 & 0.357 & 0.357 & 0.120 \\

BEVDepth~\cite{li2022bevdepth}
& VoV99 & $1600\times640$ & $90^{\ddagger}$
& 60.0 & 50.3
& 0.445 & 0.245 & 0.378 & 0.320 & 0.126 \\

HoP-BEVFormer~\cite{zong2023temporal}
& VoV99 & $1600\times640$ & 24
& 60.3 & 51.7
& 0.501 & 0.245 & 0.346 & 0.362 & 0.105 \\

BEVStereo~\cite{li2023bevstereo}
& VoV99 & $1600\times640$ & $90^{\ddagger}$
& 61.0 & 52.5
& 0.431 & 0.246 & 0.358 & 0.357 & 0.138 \\

CAPE-T~\cite{xiong2023cape}
& VoV99 & $1600\times640$ & $24^{\ddagger}+60$
& 61.0 & 52.5
& 0.503 & 0.242 & 0.361 & 0.306 & 0.114 \\

FB-BEV~\cite{li2023fb}
& VoV99 & $1600\times640$ & 30
& 62.4 & 53.7
& 0.439 & 0.250 & 0.358 & 0.270 & 0.128 \\

StreamPETR~\cite{wang2023exploring}
& VoV99 & $1600\times640$ & 60
& 63.6 & 55.0
& 0.479 & 0.239 & 0.317 & 0.241 & 0.119 \\

HENet~\cite{xia2024henet}
& VoV99 \& ResNet50
& $1152\times640$ \& $704\times256$
& $54^{\ddagger}$
& 63.8 & 57.5
& 0.432 & 0.242 & 0.368 & 0.320 & 0.129 \\

SparseBEV$^{\dagger}$~\cite{liu2023sparsebev}
& VoV99 & $1600\times640$ & 24
& \underline{67.5} & \underline{60.3}
& 0.425 & 0.239 & 0.311 & 0.172 & 0.116 \\

\rowcolor{gray!15}
\textbf{PQR3D}$^{\dagger}$
& VoV99 & $1600\times640$ & 24
& \textbf{68.0} & \textbf{61.0}
& 0.406 & 0.239 & 0.318 & 0.168 & 0.118 \\

\midrule

StreamPETR~\cite{wang2023exploring}
& ViT-L & $1600\times800$ & 24
& 67.6 & 62.0
& 0.470 & 0.241 & 0.258 & 0.236 & 0.134 \\

HoP-BEVDet4D-Depth$^{\dagger}$~\cite{zong2023temporal}
& ViT-L & $1600\times640$ & $36+5^{\ddagger}$
& 68.5 & 62.4
& 0.367 & 0.249 & 0.354 & 0.171 & 0.131 \\

RayDN~\cite{liu2024ray}
& ViT-L & $1600\times800$ & 24
& 68.6 & 63.1
& 0.437 & 0.235 & 0.283 & 0.220 & 0.120 \\

RoPETR~\cite{ji2025ropetr}
& ViT-L & $1600\times640$ & 24
& 69.0 & 61.9
& 0.396 & 0.254 & 0.249 & 0.163 & 0.129 \\

SparseBEV$^{\dagger}$~\cite{liu2023sparsebev}
& ViT-L & $1600\times640$ & 24
& 70.2 & 63.5
& 0.402 & 0.235 & 0.237 & 0.154 & 0.129 \\

RoPETR-e$^{*}$~\cite{ji2025ropetr}
& ViT-L & $1600\times900$ & 24
& \underline{70.9} & \underline{64.8}
& 0.379 & 0.227 & 0.248 & 0.173 & 0.125 \\

HENet++$^{\dagger}$~\cite{xia2025henet}
& ViT-L \& VoV99 & $1600\times640$ & $54^{\ddagger}$
& 70.7 & 64.5
& 0.402 & 0.235 & 0.237 & 0.155 & 0.129 \\

\rowcolor{gray!15}
\textbf{PQR3D}$^{\dagger}$
& ViT-L & $1600\times640$ & 24
& \textbf{71.6} & \textbf{64.9}
& 0.383 & 0.235 & 0.210 & 0.142 & 0.115 \\

\bottomrule
\end{tabular}
\end{adjustbox}
\end{table*}

 


\begin{table}[t]
\centering
\caption{
Ablation studies on the nuScenes validation split using ResNet-50,
$704\times256$ input resolution, and 24 training epochs.
(a) Effect of each component. PQR: progressive query refinement;
MSA: masked self-attention; SDAE: stage-decoupled anchor embedding;
BF: BEV flip.
(b) Comparison of temporal propagation strategies under the same detector and training settings. Vanilla uses no query propagation, whereas Sequential uses a StreamPETR-style strategy that carries cached queries from one frame to the next along the video sequence.
}
\label{tab:ablation1}
\vspace{-5pt}

\small
\begin{minipage}[t]{0.52\textwidth}
\centering
\subcaption*{(a) Component ablation}
\begin{tabular}{cccccc}
\toprule
PQR & MSA & SDAE & BF
& NDS$\uparrow$ & mAP$\uparrow$ \\
\midrule
\xmark & \xmark & \xmark & \xmark
& 55.5 & 45.3 \\

\cmark & \xmark & \xmark & \xmark
& 56.6 & 46.5 \\

\cmark & \cmark & \xmark & \xmark
& 57.0 & 47.2 \\

\cmark & \cmark & \cmark & \xmark
& 57.7 & 48.2 \\

\rowcolor{gray!15}
\cmark & \cmark & \cmark & \cmark
& \textbf{58.2} & \textbf{48.6} \\
\bottomrule
\end{tabular}
\end{minipage}%
\hfill
\begin{minipage}[t]{0.44\textwidth}
\centering
\subcaption*{(b) Propagation strategy}
\begin{tabular}{lcc}
\toprule
Strategy & NDS$\uparrow$ & mAP$\uparrow$ \\
\midrule
Vanilla
& 55.5 & 45.3 \\

Sequential
& 55.7 & 45.7 \\

\rowcolor{gray!15}
\textbf{PQR (ours)}
& \textbf{56.6} & \textbf{46.5} \\
\bottomrule
\end{tabular}
\end{minipage}

\vspace{-10pt}
\end{table}

\subsection{Ablation Study}



\textbf{Component ablation.} As shown in Table~\ref{tab:ablation1}(a), we evaluate the contribution of each proposed component by incrementally adding it to the baseline. Progressive Query Refinement (PQR) provides the largest single improvement of $+1.1$ NDS and $+1.2$ mAP, demonstrating that progressively aligning and refining high-confidence queries toward the target frame effectively leverages information from earlier frames.
Adding masked self-attention (MSA) yields a further $+0.4$ NDS and $+0.7$ mAP by isolating denoising queries while preserving interactions between regular and propagated queries. 
Stage-decoupled anchor embedding (SDAE) brings an additional $+0.7$ NDS and $+1.0$ mAP by applying the position embedding before self-attention and injecting the remaining anchor attributes afterward, reducing conflicts among temporal queries.
Finally, BEV flip augmentation (BF) brings $+0.5$ NDS and $+0.4$ mAP, providing spatial regularization compatible with temporal propagation. Overall, the four components improve the baseline by $+2.7$ NDS and $+3.3$ mAP, with complementary gains from each component.

\textbf{Propagation strategy.}
As shown in Table~\ref{tab:ablation1}(b), we compare three propagation strategies under the same settings. Vanilla uses no query propagation, whereas the StreamPETR-style Sequential baseline uses sequential video sampling and carries cached queries from one frame to the next, yielding only $+0.2$ NDS and $+0.4$ mAP.
 In contrast, PQR independently processes each temporal window and progressively refines queries from earlier frames toward the target frame, improving Vanilla by $+1.1$ NDS and $+1.2$ mAP. PQR also outperforms Sequential by $+0.9$ NDS and $+0.8$ mAP, demonstrating the advantage of window-based progressive refinement over frame-by-frame propagation.

\begin{figure*}[t!]
\begin{center}
\includegraphics[width=1.0\textwidth]{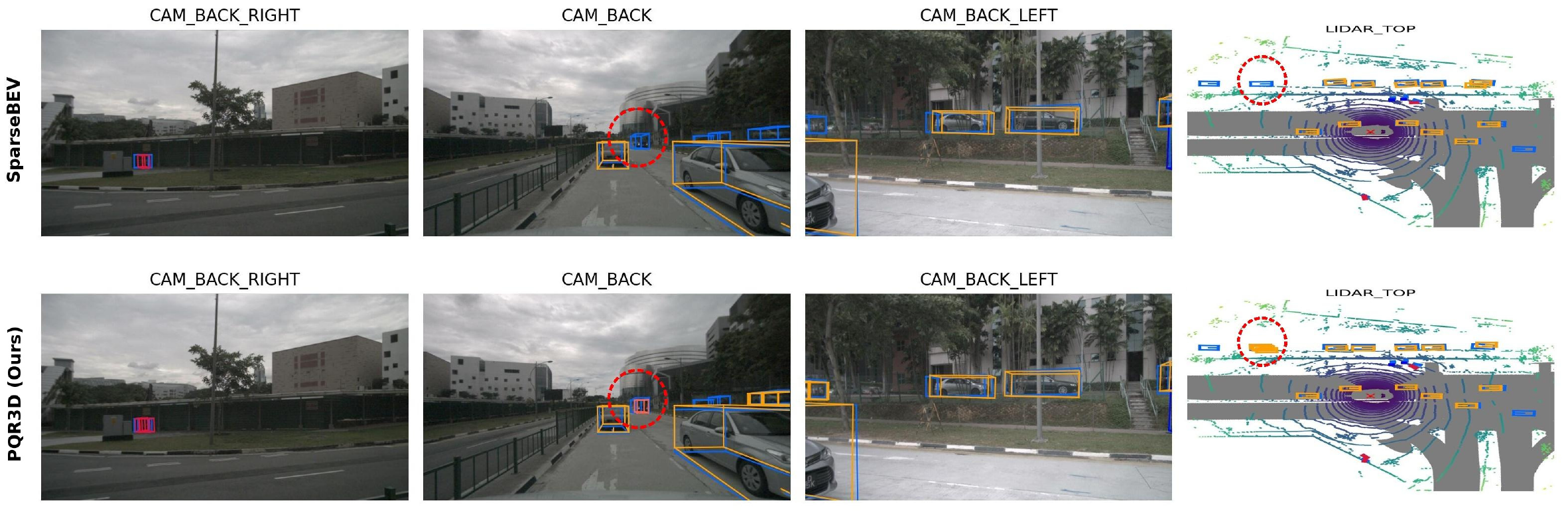}
\end{center}
\vspace{-10pt}
\caption{Qualitative comparison of SparseBEV (top) and PQR3D (bottom) on the nuScenes val set. Ground-truth boxes are shown in blue and predictions are class-colored (orange for car, crimson for bicycle, etc.). Red dashed circles mark regions of interest.}
\label{fig:visualization}
 \vspace{-10pt}
\end{figure*}

\subsection{Visualization}
Figure~\ref{fig:visualization} shows a qualitative comparison between the baseline SparseBEV and our PQR3D on the nuScenes validation set. Within the highlighted region marked by the red dashed circle, SparseBEV fails to detect a distant vehicle, whereas PQR3D successfully recovers it by progressively refining object queries with temporal evidence from preceding frames. This improvement is visible in both the rear-camera view and the LiDAR BEV visualization, demonstrating that progressive query refinement helps recover objects with weak visual evidence in the current frame.




\section{Conclusion}
\label{sec:conclusion}
We present PQR3D, a progressive query
refinement framework for multi-view 3D object detection. Within each reference-conditioned temporal window, PQR3D progressively refines object queries from earlier keyframes toward the target frame using ego-pose and velocity compensation. A masked self-attention mechanism is designed for the interactions among denoising, regular, and propagated queries to prevent information leakage while allowing regular and propagated queries to attend to each other. Furthermore, a stage-decoupled anchor embedding injects geometric and motion attributes without introducing conflicts during attention computation. Extensive experiments on nuScenes demonstrate that PQR3D achieves state-of-the-art performance across multiple backbone configurations. Future work includes extending PQR3D to broader 3D perception tasks such as BEV segmentation and occupancy prediction.

\subsubsection*{Acknowledgments}

Research was sponsored by the Army Research Laboratory and was accomplished under Cooperative Agreement Number W911NF-23-2-0224. The views and conclusions contained in this document are those of the authors and should not be interpreted as representing the official policies, either expressed or implied, of the Army Research Laboratory or the U.S. Government. The U.S. Government is authorized to reproduce and distribute reprints for Government purposes notwithstanding any copyright notation herein.

\section*{AI use statement}

Generative AI tools were used for language editing, including grammar and
spelling correction, word choice, and readability improvement; to assist in
reviewing source code for potential errors; and to assist in creating
Figs.~\ref{fig:propagation}, \ref{fig:sdae}(a),
\ref{fig:visualization}, \ref{fig:fps} and~\ref{fig:visualization_13}.
Figs.~\ref{fig:architecture} and \ref{fig:sdae}(b) were created entirely
by the authors. All AI-assisted text, code-review suggestions, and visual
materials were reviewed, revised, and verified by the authors. All research
ideas and methods were developed by the authors, who also designed and
conducted the experiments, analyzed the results, and drew the conclusions.
The authors take full responsibility for the final content of this paper.

\section*{Ethics statement}

This work uses the publicly available nuScenes dataset and collects no new human-subject data. Improved 3D object detection may enhance autonomous-driving safety, but perception failures could lead to unsafe decisions. PQR3D is a research prototype, and real-world deployment requires extensive validation, system redundancy, and fail-safe mechanisms.

\section*{Reproducibility statement}

Implementation and training details are provided in Sec.~\ref{subsec:implementation_details} and Appendix~\ref{app:implementation}. The source code, configuration files, environment setup instructions, training and evaluation commands, trained checkpoints, and training logs are publicly available at \url{https://github.com/huiyegit/PQR3D}. These resources support reproduction of the reported nuScenes experiments and results.



\bibliography{iclr2027_conference}
\bibliographystyle{iclr2027_conference}

\clearpage
\appendix

\section{Additional Experiments and Analysis}
\label{app:additional_results}

\subsection{Implementation Details}
\label{app:implementation}

Unless otherwise specified, models used for benchmark comparison and
ablation studies are trained for 24 epochs using
AdamW~\cite{loshchilov2019decoupled} with a batch size of 8, a base
learning rate of $2\times10^{-4}$, and cosine
annealing~\cite{loshchilov2017sgdr}. The Transformer decoder contains
$L=6$ layers.

The validation configurations use the current frame and eight preceding
frames sampled at approximately $0.5$-s intervals. Test configurations
marked with $\dagger$ additionally use future frames and therefore
correspond to offline rather than causal online inference.

We use focal loss~\cite{lin2017focal} for classification and L1 loss for
bounding-box regression. For query denoising, we set $G=10$ and use a
noise scale of $\lambda=0.4$, following
SparseBEV~\cite{liu2023sparsebev}. Unless otherwise specified, PQR3D uses
two refinement steps with $k=256$ propagated queries per step.

ResNet-50 and ResNet-101 models are trained on $8\times$ NVIDIA RTX 4090
GPUs, while VoV99 and ViT-L models are trained on $8\times$ NVIDIA A100
GPUs. The lightweight 625-query configuration used for the efficiency
comparison is trained for 48 epochs.

\FloatBarrier

\subsection{Inference Speed}
\label{app:inference_speed}

\begin{figure}[H]
    \centering
    \includegraphics[width=0.72\linewidth]{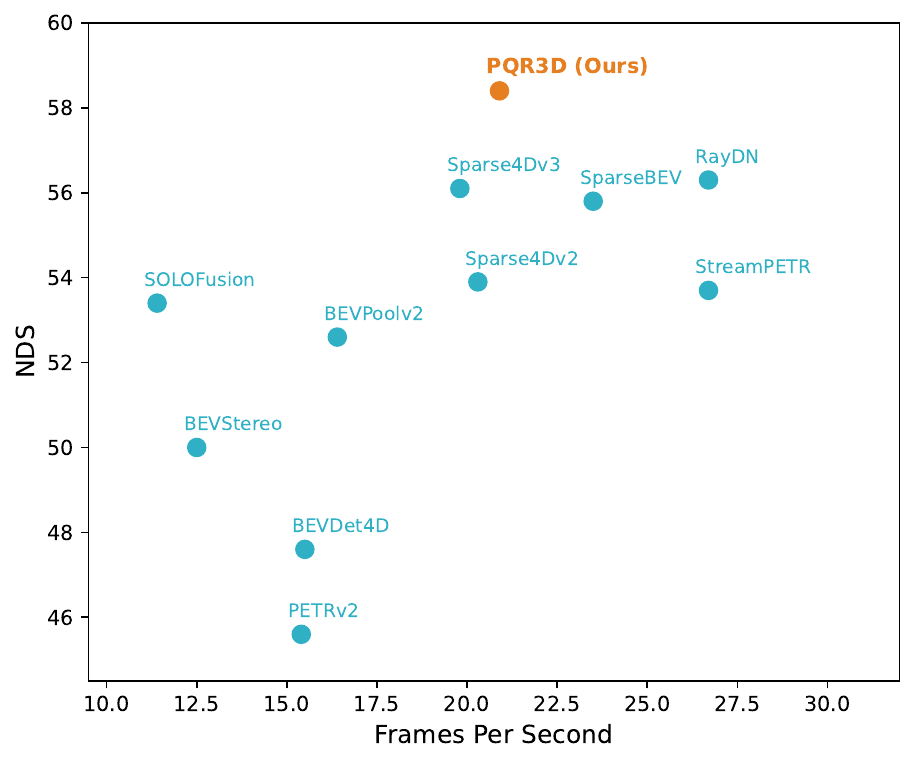}
    \caption{Speed--accuracy comparison of PQR3D with existing methods
    on the nuScenes validation set in terms of frames per second (FPS)
    and NDS.}
    \label{fig:fps}
\end{figure}

The standard PQR3D configuration uses 900 queries and is trained for
24 epochs. For the efficiency comparison, we additionally evaluate a
lightweight configuration with 625 queries trained for 48 epochs. As
shown in Figure~\ref{fig:fps}, all methods are evaluated on a single
NVIDIA RTX 3090 GPU under the same inference settings. The lightweight
PQR3D achieves $58.5$ NDS and $48.1$ mAP while running at approximately
$20$ FPS. At comparable throughput, it outperforms Sparse4Dv3 by $2.4$
NDS points. Although RayDN and SparseBEV run faster, PQR3D improves NDS
by $2.2$ and $2.7$ points, respectively. These results demonstrate that
PQR3D provides a favorable balance between detection accuracy and
inference speed.

\FloatBarrier

\subsection{Additional Ablation Studies}
\label{app:additional_ablations}

We further study the number of progressive refinement steps, the number
of propagated queries, and the anchor embedding design.

\begin{table}[H]
    \centering
    \caption{Effect of the number of progressive refinement steps on
    detection accuracy and GPU memory usage.}
    \label{tab:refinement_steps}
    \small
    \setlength{\tabcolsep}{8pt}
    \renewcommand{\arraystretch}{1.15}
    \begin{tabular}{cccc}
        \toprule
        Steps
        & NDS$\uparrow$
        & mAP$\uparrow$
        & Mem.\ (GB)$\downarrow$ \\
        \midrule
        0 & 55.5 & 45.3 & 8.1 \\
        1 & 56.2 & 46.1 & 8.3 \\
        \rowcolor{gray!15}
        2 & 56.6 & \textbf{46.5} & 8.7 \\
        3 & \textbf{56.8} & 46.4 & 9.3 \\
        \bottomrule
    \end{tabular}
\end{table}

\textbf{Number of refinement steps.}
As shown in Table~\ref{tab:refinement_steps}, one progressive refinement
step improves the baseline by $+0.7$ NDS and $+0.8$ mAP. Using two steps
provides a further gain of $+0.4$ NDS and $+0.4$ mAP by incorporating an
additional historical keyframe. A third step yields only $+0.2$ NDS,
decreases mAP by $0.1$, and increases memory usage from $8.7$ to
$9.3$ GB, indicating diminishing returns from additional refinement.
We therefore use two refinement steps by default to balance detection
accuracy and memory usage.

\begin{table}[H]
    \centering
    \caption{Ablation studies on propagated query count and anchor
    embedding design. (a) Number of propagated queries $k$.
    (b) Anchor embedding design. Pos: position-only embedding.}
    \label{tab:ablation2}
    \small

    \begin{minipage}[t]{0.48\textwidth}
        \centering
        \subcaption*{(a) Propagated query count $k$}
        \setlength{\tabcolsep}{5pt}
        \renewcommand{\arraystretch}{1.1}
        \begin{tabular}{cccc}
            \toprule
            $k$
            & NDS$\uparrow$
            & mAP$\uparrow$
            & Mem.\ (GB)$\downarrow$ \\
            \midrule
            0   & 55.5 & 45.3 & 8.1 \\
            128 & 56.3 & 46.0 & 8.6 \\
            \rowcolor{gray!15}
            256 & 56.6 & \textbf{46.5} & 8.7 \\
            512 & \textbf{56.7} & 46.2 & 8.9 \\
            \bottomrule
        \end{tabular}
    \end{minipage}%
    \hfill
    \begin{minipage}[t]{0.48\textwidth}
        \centering
        \subcaption*{(b) Anchor embedding design}
        \setlength{\tabcolsep}{7pt}
        \renewcommand{\arraystretch}{1.1}
        \begin{tabular}{lcc}
            \toprule
            Variant & NDS$\uparrow$ & mAP$\uparrow$ \\
            \midrule
            Pos & 57.0 & 47.2 \\
            Coupled & 57.2 & 46.9 \\
            \rowcolor{gray!15}
            Stage-decoupled
            & \textbf{57.7}
            & \textbf{48.2} \\
            \bottomrule
        \end{tabular}
    \end{minipage}
\end{table}

\textbf{Propagated query count.}
As shown in Table~\ref{tab:ablation2}(a), setting $k=128$ improves the
baseline by $+0.8$ NDS and $+0.7$ mAP, showing that even a small set of
propagated queries is beneficial. Increasing $k$ to $256$ provides
further gains of $+0.3$ NDS and $+0.5$ mAP by transferring more
high-confidence object queries between refinement steps. Increasing
$k$ to $512$, however, improves NDS by only $0.1$, decreases mAP by
$0.3$, and adds $0.2$ GB of memory. We therefore use $k=256$ by default.

\textbf{Anchor embedding design.}
As shown in Table~\ref{tab:ablation2}(b), the coupled design improves
NDS by $0.2$ but decreases mAP by $0.3$ compared with the position-only
baseline. This suggests that injecting all anchor attributes before
self-attention can introduce conflicts between regular and propagated
queries. In contrast, our stage-decoupled design injects position before
self-attention and adds size, orientation, and velocity afterward. It
achieves $57.7$ NDS and $48.2$ mAP, improving the position-only baseline
by $0.7$ NDS and $1.0$ mAP. These results show that stage-wise attribute
injection enriches query representations while reducing conflicts among
temporal queries.

\FloatBarrier

\subsection{Additional Qualitative Results}
\label{app:qualitative_results}

\begin{figure}[H]
    \centering
    \includegraphics[width=\textwidth]
    {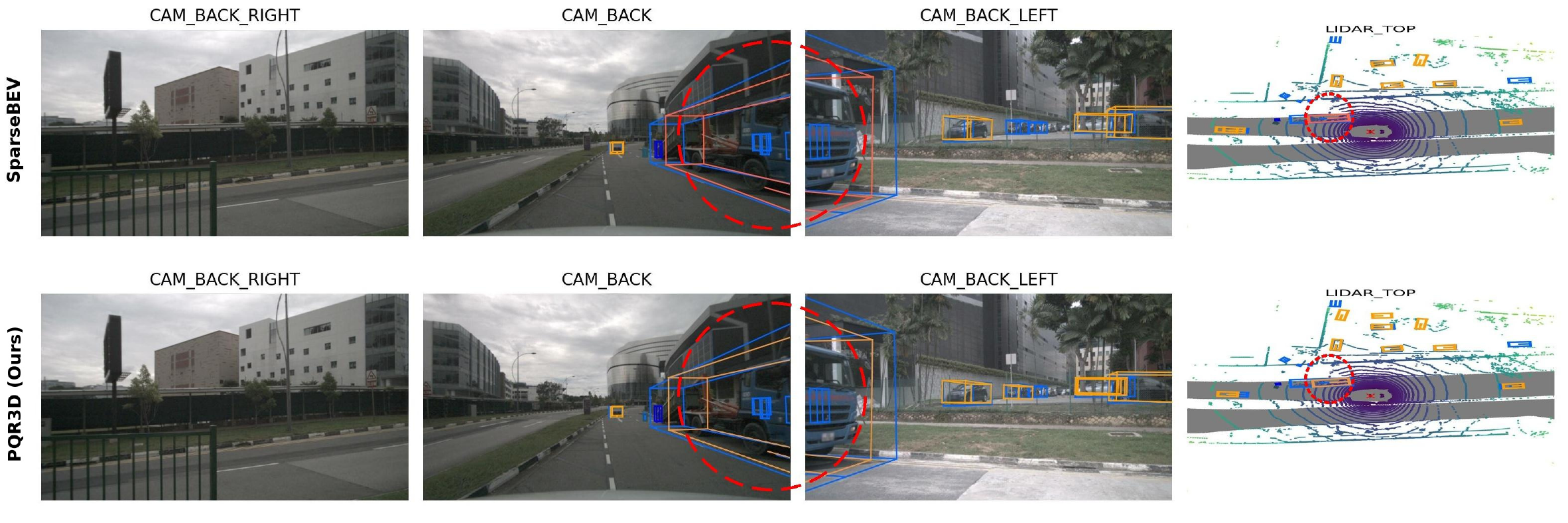}
    \caption{Qualitative comparison of SparseBEV (top) and PQR3D
    (bottom) on the nuScenes validation set. Ground-truth boxes are
    shown in blue, and predictions are class-colored. Red dashed
    circles indicate regions of interest.}
    \label{fig:visualization_13}
\end{figure}

Figure~\ref{fig:visualization_13} presents another qualitative comparison
on the nuScenes validation set. In the highlighted region, SparseBEV
produces a poorly localized bounding box for a partially occluded truck,
whereas PQR3D provides more accurate localization through progressive
query refinement within the temporal window. This example shows that
progressive query refinement improves both object recall and localization
accuracy under occlusion.

\FloatBarrier

\end{document}

%% file: math_commands.tex
\usepackage{amsmath,amsfonts,bm}

\def\eqref#1{equation~\ref{#1}}

\def\1{\bm{1}}

\DeclareMathAlphabet{\mathsfit}{\encodingdefault}{\sfdefault}{m}{sl}
\SetMathAlphabet{\mathsfit}{bold}{\encodingdefault}{\sfdefault}{bx}{n}

